\documentclass[final]{cvpr}

\usepackage{times}
\usepackage{epsfig}
\usepackage{graphicx}
\usepackage{amsmath}
\usepackage{amssymb}
\usepackage{bm}
\usepackage{array}
\usepackage{multirow}
\usepackage{float}
\usepackage{booktabs}

\newcommand{\LAB}{\text{L}^{*}\text{a}^{*}\text{b}^{*}}
\newcommand{\Lcol}{\text{L}^{*}}
\newcommand{\ABcol}{\text{a}^{*}\text{b}^{*}}

\usepackage[pagebackref=true,breaklinks=true,colorlinks,bookmarks=false]{hyperref}

\def\cvprPaperID{35} %
\def\confYear{EarthVision 2021}

\begin{document}

\title{Self-Supervised Learning of Remote Sensing Scene Representations Using Contrastive Multiview Coding}

\author{Vladan Stojnić \qquad Vladimir Risojević \vspace{0.5em} \\
Faculty of Electrical Engineering, University of Banja Luka, Bosnia and Herzegovina\\
\{{\tt\small vladan.stojnic, vladimir.risojevic}\}{\tt\small @etf.unibl.org}
}

\maketitle

\thispagestyle{empty}

\begin{abstract}
   In recent years self-supervised learning has emerged as a promising candidate for unsupervised representation learning. In the visual domain its applications are mostly studied in the context of images of natural scenes. However, its applicability is especially interesting in specific areas, like remote sensing and medicine, where it is hard to obtain huge amounts of labeled data. In this work, we conduct an extensive analysis of the applicability of self-supervised learning in remote sensing image classification. We analyze the influence of the number and domain of images used for self-supervised pre-training on the performance on downstream tasks. We show that, for the downstream task of remote sensing image classification, using self-supervised pre-training on remote sensing images can give better results than using supervised pre-training on images of natural scenes. Besides, we also show that self-supervised pre-training can be easily extended to multispectral images producing even better results on our downstream tasks.
\end{abstract}

\section{Introduction}

The advent of deep learning based methods enabled rapid progress in computer vision, as can be seen by the results on many benchmark datasets like ImageNet~\cite{russakovsky2015imagenet} and MS-COCO~\cite{lin2014microsoft}. However, most of these improvements came from the usage of supervised learning based methods that need huge amounts of labeled training data. This need for large labeled datasets poses a significant problem as creating them is a highly time-consuming and error-prone task. This is especially visible in highly specialized areas, like remote sensing and medicine, where data labeling can only be done by domain experts.

Over the years transfer learning emerged as a standard technique for overcoming this need for big labeled datasets. It is based on the idea of using neural networks pre-trained on already available labeled datasets, like ImageNet, as either feature extractors or for finetuning on small datasets. This allowed usage of these supervised deep learning based methods for small datasets, but some limitations remained. One major limitation in applying these techniques for remote sensing images is that most remote sensing images contain information outside of the visible spectrum, while the neural networks are usually pre-trained on RGB images as most of the big labeled datasets contain only RGB images. This forces us to drop all information outside of the visible spectrum if we want to use a neural network pre-trained on RGB images, which significantly limits the amount of information we use in the analysis. Even if remote sensing images only contain RGB color channels, a possible drawback of using models pre-trained on ImageNet is that images of natural scenes present in ImageNet look quite different from remote sensing images, especially if these are satellite images. Besides, different missions like Sentinel, Landsat, and others provide us with an ever-growing amount of unlabeled multispectral images which makes it even more interesting to study the possibilities of learning visual representations in an unsupervised manner.

Recently, self-supervised learning~\cite{kolesnikov2019revisiting,goyal2019scaling,jaiswal2021survey} emerged as a promising candidate for overcoming the need for large labeled datasets in representation learning. It is based on the idea of training a neural network to solve the pretext task for which labels can be obtained directly from the data. This pretext task should be complex enough to allow the neural network to learn meaningful image representations. A neural network trained in a self-supervised fashion can then be used for transfer learning in the same way as we would use neural networks pre-trained on supervised tasks. However, as this is an unsupervised technique of pre-training it allows us to take advantage of a large number of available unlabeled images. This possibility of using unlabeled images for representation learning is especially interesting in areas where we have multispectral images or images that simply look differently than images of natural scenes. In these areas using in-domain images for pre-training could allow us to overcome the issues that we have with classical transfer learning.

Applications of self-supervised learning, in working with images of natural scenes, have been studied quite well in the last few years and they show very promising results. In the last year, self-supervised learning provided representations that transfer to classification, segmentation, and detection tasks even better than representations learned by supervised learning~\cite{chen2020simple,caron2020unsupervised}. However, the application of self-supervised learning methods in remote sensing has not been studied a lot. Most of these applications either used small amounts, up to 50,000 images, of unlabeled training data~\cite{stojnic2018evaluation,stojnic2018analysis,tao2020remote} or tested learned representations on a small number of datasets~\cite{ayush2020geography,kang2020deep,vincenzi2020color}.

For these reasons, in this work we give a detailed analysis of the applicability of self-supervised learning in remote sensing. First, we analyze the influence of the number and domain of images used for self-supervised pre-training on the performance on downstream tasks. Second, we compare representations learned by self-supervised learning to the representations learned by supervised learning. Third, we analyze different strategies for expanding a self-supervised algorithm proposed for RGB images to the case of multispectral images. Fourth, we analyze the possibility of finetuning these self-supervised models and the influence it has on results on downstream tasks. Finetuning is usually skipped in most self-supervised learning papers and these models are usually only used as feature extractors. Besides that, we compared these results to the results obtained by finetuning supervisedly trained models.

We show that, for the classification of remote sensing images, using self-supervised pre-training on remote sensing images can give better results than using supervised pre-training on images of natural scenes, even when we use significantly smaller number of images. We also apply self-supervised pre-training on multispectral images and show that it further improves the results on downstream tasks. Code and trained models used in this paper are available at {\footnotesize\url{https://github.com/vladan-stojnic/CMC-RSSR}}.

\section{Related Work}
\label{sec:related}

\subsection{Self-supervised learning of visual representations}

Self-supervised learning of visual representations is a fast-growing subfield of unsupervised learning and, over the years, many methods have been proposed and applied. All the proposed methods can be categorized into three groups depending on the type of pretext task they use.

In the first group, we have methods based on the pretext task of image reconstruction or solving some natural tasks. In these pretext tasks, self-supervised models are trained to remove a degradation artificially added to images or to solve some simple natural tasks in image understanding. For example, in~\cite{doersch2015unsupervised} a pretext task is formulated in a way that a model needs to predict how one block of the image is positioned compared to another block of the same image. A similar pretext task is presented in~\cite{noroozi2016unsupervised} where a model is trained to solve a jigsaw puzzle consisting of nine shuffled image blocks. In~\cite{pathak2016context} the authors propose a pretext task based on solving the image inpainting problem. Another interesting pretext task is proposed in~\cite{gidaris2018unsupervised} where a model learns to predict the natural orientation of the image. A pretext task based on colorizing an image is presented in~\cite{larsson2017colorization,zhang2016colorful} while in~\cite{zhang2017split} authors propose a pretext task where they train a model to predict both the colors of the image from its luminance as well as predict the luminance from its colors. The methods from this group perform generally worse than the methods from the other two groups. Moreover, some of these methods, such as image inpainting, as well as predicting natural orientations or relative positions of image blocks cannot be unambiguously defined for remote sensing images, which makes them inapplicable to this domain.

The second group of methods for construction of pretext tasks is based on the idea of predicting synthetically generated labels. In this case self-supervised models solve a classical classification problem, but instead of predicting human supplied labels they predict labels that are automatically generated. In~\cite{dosovitskiy2015discriminative} every image is its own class and other samples from that class are generated by applying different augmentations to the original image. Another way of generating labels is proposed in~\cite{caron2018deep}. This time classes are determined by applying k-means clustering to the representations extracted from a self-supervised model. New classes are determined at the start of each epoch, while each image is assigned a label that corresponds to the index of the closest cluster.

\begin{table*}[ht]
    \centering
    \begin{tabular}{cccc}
         \toprule
         Dataset & Number of images & Image sizes & Spatial resolution (m)  \\
         \midrule
         ImageNet       & 1,281,167 &  up to 4488x7056 & - \\
         NWPU-RESISC45  & 31,500   &  256x256 & 0.2 - 30  \\
         DOTA           & 2,806    &  800x800 - 4000x4000 & not specified  \\
         NWPU VHR-10    & 800     &  381x601 - 1028x1728 & 0.08 - 2 \\
         BigEarthNet    & 269,695  &  up to 120x120 & 10 - 60 \\
         \bottomrule
    \end{tabular}
    \vspace{0.3em}
    \caption{Details of datasets used for self-supervised learning. (Number of images denotes number of images used for self-supervised learning and not the number of images in a dataset.)}
    \label{tab:ssl_datasets}
\end{table*}

The third group of self-supervised methods is based on the idea of contrastive learning. In the case of contrastive learning, the pretext task is defined in a way that representations of similar (positive) images should be close in the representation space, while the representations of dissimilar (negative) images should be as far as possible. Now the main task is how to find similar/dissimilar images. In~\cite{oord2018representation,henaff2020data} the authors propose to use patches extracted from the same image as positive pairs and patches from different images as negative pairs. Another method for finding positive pairs is based on the idea of applying different augmentations to the same image and using them as positive pairs~\cite{chen2020simple,wu2018unsupervised,ye2019unsupervised,he2020momentum}, while the negative pairs again come from different images. A method based on using representations from different neural network layers as positive pairs is proposed in~\cite{hjelm2018learning,bachman2019learning}. In~\cite{tian2019contrastive} the authors propose to use disjoint subsets of image channels as positive pairs while they use subsets of channels coming from different images as negative pairs. Another method is proposed in~\cite{caron2020unsupervised,li2020prototypical}. This time authors propose to contrast image representations with their cluster assignments. The process is quite similar to the previously mentioned methods that use clustering for generating synthetic image labels. They calculate representations of positive/negative images but instead of directly contrasting representations, they cluster the representations and contrast them with cluster assignments of their positive/negative pairs.

\subsection{Self-supervised learning in remote sensing}

Lately, some research has been done in the area of applying self-supervised learning to the analysis of remote sensing images. In~\cite{stojnic2018evaluation,stojnic2018analysis} the authors analyzed the possibilities of using split-brain autoencoder~\cite{zhang2017split} in the analysis of aerial images. They analyzed the influence of the number of images used for self-supervised learning, as well as usage of different color channels, on the results obtained on the downstream task of aerial image classification. A method that is quite similar to contrastive learning with different image augmentations is proposed in~\cite{ayush2020geography}. However, instead of using different augmentations of the same image as a positive pair, they use satellite recordings of the same geospatial areas in different time frames. Also, besides the contrastive loss, they introduce a loss term that is based on the classification of the image geolocation. Classification of a geolocation is based on the idea of predicting artificial labels. Artificial labels are generated by k-means clustering of latitude and longitude coordinates of every image in a dataset. In~\cite{tao2020remote} the authors analyze possibilities of using different self-supervised methods, specifically the ones based on image inpainting~\cite{pathak2016context}, context prediction~\cite{doersch2015unsupervised}, and contrastive learning using different augmentations. However, they only train these methods on small datasets of remote sensing images ranging from 20,000 to 30,000 images. A method proposed in~\cite{kang2020deep} is based on contrastive learning with different augmentations. They train their self-supervised models on relatively big datasets with 100,000 images, but they only test them on two downstream tasks NAIP~\cite{jean2019tile2vec} and EuroSAT~\cite{helber2019eurosat}. In~\cite{vincenzi2020color} the authors propose a method similar to the image colorization, but they train a model to predict information from the visible part of the spectrum from the information outside of the visible spectrum. The authors trained the model using the BigEarthNet dataset~\cite{sumbul2019bigearthnet,sumbul2020bigearthnet} that is quite big with 269,695 training images, but they again only tested learned representation on two downstream tasks.

In this work, we train the self-supervised algorithm on three large datasets of: (a) images of natural scenes, (b) high-resolution remote sensing images, and (c) low-resolution remote sensing images. In addition, we apply it to both RGB and multispectral remote sensing images. Finally, we evaluate learned representations on four downstream tasks of both single-label and multi-label classification problems.

\begin{table*}[ht]
    \centering
    \begin{tabular}{cccccc}
         \toprule
         Dataset & Annotation type & Number of categories & Number of images & Image sizes & Spatial resolution (m)  \\
         \midrule
         AID       & single-label & 30 & 10,000 &  600x600 & 0.5 - 8 \\
         MLRSNet  & multi-label & 60 & 109,161   &  256x256 & 0.1 - 10  \\
         BigEarthNet & multi-label & 19 & 519,284    &  up to 120x120 & 10 - 60  \\
         So2Sat & single-label & 17    & 376,485     &  32x32 & 10 \\
         \bottomrule
    \end{tabular}
    \vspace{0.3em}
    \caption{Details of datasets used for the evaluation of self-supervised representations.}
    \label{tab:testing_datasets}
\end{table*}

\section{Datasets}

To perform the experiments in this paper we use two groups of datasets. One group contains datasets used for training of the self-supervised algorithm and the second group contains datasets used for the evaluation of the representations obtained using self-supervised pre-training.

\subsection{Datasets used for self-supervised pre-training}

To train self-supervised models we use three different datasets. One dataset consists of images of natural scenes and will be used as a baseline, while the two remaining datasets consist of remote sensing images. One of these datasets contains high-resolution images, while the other one contains low-resolution images. It is important to note that even though most of these datasets contain labels we do not use them in any way for training of our self-supervised models. Details about these datasets can be seen in Table~\ref{tab:ssl_datasets}.

For the dataset of natural scenes, we chose ImageNet~\cite{russakovsky2015imagenet} as it is a standard dataset used for training both supervised and self-supervised models in the literature. In this work, we use only the training set of 1.2M images from ImageNet1000 dataset.

As the high-resolution remote sensing dataset, we use the dataset created by combining images from NWPU-RESISC45~\cite{cheng2017remote}, DOTA~\cite{xia2018dota}, and NWPU VHR-10~\cite{su2019object} datasets. Some of the images in DOTA dataset contain a black border, so we manually extracted regions without it from every image in the dataset. From these regions, we then extracted non-overlapping patches of 256x256 pixels that we used as samples in our dataset for self-supervised learning. For NWPU VHR-10 dataset, we only extracted non-overlapping patches of 256x256 pixels from the original images and use these patches as samples in our dataset. Since NWPU-RESISC45 dataset originally consists of 256x256 pixel images we use these original images as samples in our dataset. Using this procedure we created a dataset of 196,215 images that we use for training of our self-supervised algorithm. We call this dataset NWPU-DOTA throughout this paper.

For the low-resolution dataset, we use images from  BigEarthNet~\cite{sumbul2019bigearthnet,sumbul2020bigearthnet} dataset. This dataset consists of 519,284 multispectral images collected by Sentinel-2 satellite, but we only use images from the training set of BigEarthNet, 269,695 images in total, as the dataset for training of the self-supervised algorithm.

\subsection{Datasets used for the evaluation of learned representations}

To evaluate the representations learned by the self-supervised algorithm we use four remote sensing datasets: two with high-resolution and two with low-resolution images. Details about these datasets can be seen in Table~\ref{tab:testing_datasets}.

AID~\cite{xia2017aid} is a high-resolution scene classification dataset of aerial images. Every image in this dataset is labeled with one category from the set of 30 predefined categories. On this dataset, we use 50\% - 50\% training/test split for training and evaluation of our models, as previously used in literature.

MLRSNet~\cite{qi2020mlrsnet} is a dataset of high-resolution aerial images. This dataset was developed for the problem of scene classification in a multi-label setting. Every image in this dataset is assigned 1 to 13 labels from the set of 60 predefined categories. For training and evaluation of our models, we use a 20\%-80\% training/test split as suggested in~\cite{qi2020mlrsnet}.

BigEarthNet is a dataset of multispectral satellite images for multi-label land use classification. There are two versions of this dataset with 45~\cite{sumbul2019bigearthnet} and 19~\cite{sumbul2020bigearthnet} predefined categories, but in this work, we only consider a version of the dataset with 19 categories. For training/validation/test split we use the same split defined in~\cite{sumbul2020bigearthnet} with 269,695, 123,723, and 125,866 images, respectively.

So2Sat~\cite{zhu2019so2sat} represents a dataset of multispectral satellite images collected by Sentinel-2 and SAR images collected by Sentinel-1 designed for local climate zones classification. Every image in this dataset is labeled with one category from the set of 17 predefined categories. In this work, we use only multispectral images and the official training/validation split with 352,366 and 24,119 images, respectively.

\section{Method}

The method we use in this work consists of two steps: (1) self-supervised pre-training and (2) transfer learning and evaluation of learned representations.

\subsection{Self-supervised pre-training}

\begin{figure*}
   \centering
   \includegraphics[]{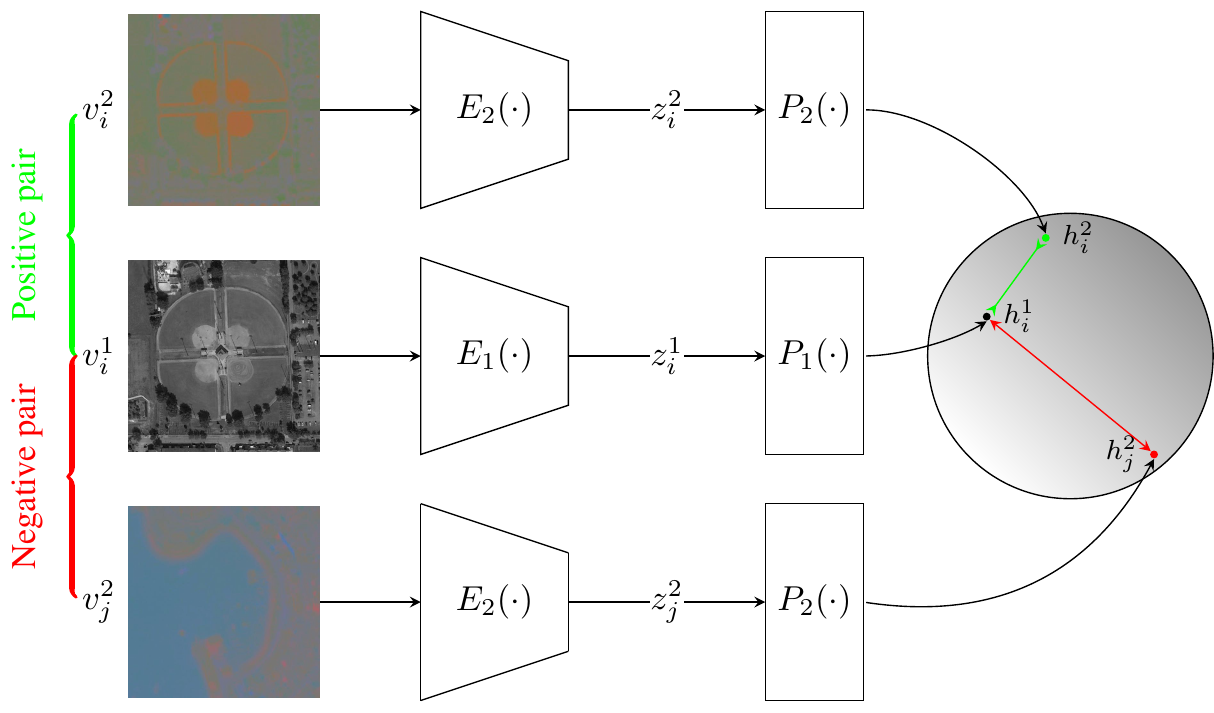}
   \caption{Schematic representation of the CMC algorithm.}
   \label{fig:cmc}
\end{figure*}

As the first step in our experiments, we use Contrastive Multiview Coding - CMC~\cite{tian2019contrastive} for self-supervised pre-training. We chose CMC because it gave good results for images of natural scenes and could be readily adapted to multispectral images. We believe that similar results would be obtained by using the majority of the methods from the second and third groups in Section~\ref{sec:related}. 

CMC is based on the idea of contrasting representations of different image views. More specifically, given a set of images $\{x_1, x_2, \dotsc, x_N\}$, each image $x_i$ is split into two views, $v_i^1$ and $v_i^2$, containing different subsets of channels. Each of these views is then processed using encoding, $z_i^1=E_1(v_i^1)$, $z_i^2=E_2(v_i^2)$, and projection, $h_i^1=P_1(z_i^1)$, $h_i^2=P_2(z_i^2)$, networks that output data and contrastive representations, respectively. The model is then trained to pull together contrastive representations of views coming from the same image, $h_i^1$ and $h_i^2$, while pushing apart contrastive representations of views from different images, $h_i^1$ and $h_j^2$, as shown in Figure~\ref{fig:cmc}. This is achieved by training the encoding and projection networks to minimize the loss
\begin{equation}
\label{eq:loss}
L^{V^1, V^2} = -\underset{\left\{ v_1^1, v_1^2, \ldots, v_{k+1}^2 \right\}}{\mathbb{E}}
    \left[ \log \frac{s \left( h_1^1, h_1^2 \right)}{\sum_{j=1}^{k+1} s \left( h_1^1, h_j^2 \right)} \right],
\end{equation}
where view $V^1$ is an anchor, and one positive and $k$ negative images are sampled from view $V^2$. Discriminating function $s\left( \cdot, \cdot \right)$ is cosine similarity of contrastive representations with temperature hyperparameter $\tau$
\begin{equation}
    s \left( h^1, h^2 \right) = \exp{ \left( \frac{h^1 \cdot h^2}{\|h^1\| \|h^2\|} \frac{1}{\tau} \right) }.
\end{equation}
Finally, the objective function for CMC is obtained by computing the losses~(\ref{eq:loss}) using both views as anchors and summing them
\begin{equation}
    L = L^{V_1, V_2} + L^{V_2, V_1}.
\end{equation}
Since $k$ in~(\ref{eq:loss}) can be very large, Noise Contrastive Estimation~\cite{gutmann2010noise} is used to approximately compute the loss and reduce computational requirements.

In this work, we use a ResNet50~\cite{he2016deep} based architecture for the encoding network of the CMC model, and for the projection network we use a fully-connected linear layer followed by L2 normalization. We train our CMC models for $400$ epochs using SGD with batch size $100$, learning rate $0.03$, momentum $0.9$, and weight decay $10^{-4}$. Learning rate is further reduced by a factor of $10$ after the $250\text{th}$, $300\text{th}$ and $350\text{th}$ epoch. We also employ data augmentation using random resized cropping and horizontal flipping. For calculating NCE based loss we use $k = 4096$ negative samples and the temperature parameter of $\tau = 0.07$.

\subsection{Transfer learning}
\label{sec:transfer}

\begin{figure*}[ht]
   \centering
   \includegraphics[width=0.95\linewidth]{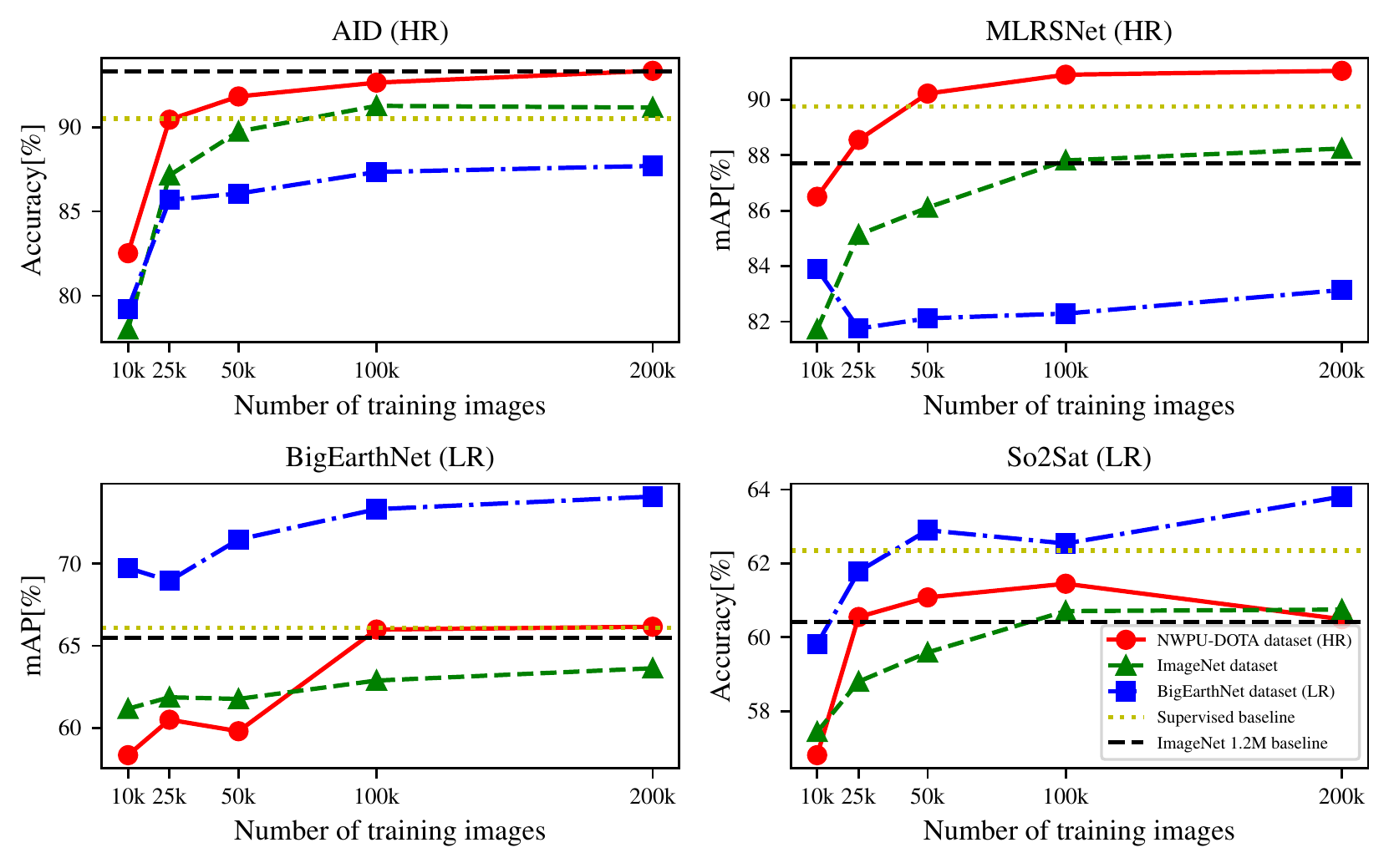}
   \caption{Results obtained using a linear classifier on top of the representations learned by supervised and self-supervised learning. Abbreviations HR and LR represent high-resolution and low-resolution remote sensing datasets, respectively.}
   \label{fig:linear}
\end{figure*}

As our main goal in this paper is to evaluate representations learned using self-supervised learning, in the next step we use previously trained CMC models for transfer learning. To conduct transfer learning, the trained CMC models are used both as feature extractors and as starting points for finetuning.

For feature extraction, we pass 224x224 pixel images through the CMC model and use concatenated outputs of both encoder networks as image representation. To evaluate these representations we train a linear classifier on top of them for our testing datasets. For single-label datasets, we use softmax activation function with categorical cross-entropy loss, while for multi-label datasets we use sigmoid activation function with binary cross-entropy loss. A linear classifier is trained for $50$ epochs using Adam optimizer with the batch size of $256$ and weight decay $10^{-2}$ for AID and MLRSNet, while for other datasets we use no weight decay. The learning rate is set to $10^{-2}$ for MLRSNet dataset and $10^{-3}$ for other datasets. This learning rate is further reduced by a factor of $5$ after the $30\text{th}$, $35\text{th}$, $40\text{th}$, and $45\text{th}$ epoch.

Similar to the feature extraction, for finetuning we put a linear layer on top of the concatenated outputs of both encoder networks and use the same activation and loss functions. This time we train the whole resulting model for $100$ epochs, without warm-up, using Adam optimizer with the batch size of $100$, learning rate of $10^{-4}$ and weight decay $10^{-4}$. Learning rate is further reduced by a factor of $5$ after the $60\text{th}$, $70\text{th}$, $80\text{th}$ and $90\text{th}$ epoch. In the process of training these finetuned models, we first rescale training images to 256x256 pixels and then extract 224x224 pixel blocks from them using random resized cropping. These blocks are also horizontally flipped with the probability of $0.5$. In the process of evaluation of the finetuned models, we pass them input images that are rescaled to 224x224 pixels.

As evaluation metrics, we use accuracy for single-label datasets and macro-averaged mean average precision (mAP) for multi-label datasets.

\section{Experimental Setup}

To assess the possibilities of CMC for learning representations of remote sensing images, we perform three experiments. First, we want to analyze the influence of the number and domain of the images used for training of the CMC model on the results obtained by training a linear classifier for our downstream tasks. To do this we train CMC models using subsets of 10,000 (10k), 25,000 (25k), 50,000 (50k), 100,000 (100k), and 196,215 (200k) images from ImageNet, NWPU-DOTA, and BigEarthNet datasets. It is important to note that in these experiments we use only RGB color channels even for datasets with multispectral images. To be precise, we train our CMC models as proposed in~\cite{tian2019contrastive}, RGB images are converted to \( \LAB \) color space and \( \Lcol \) component is used as one view of the image while \( \ABcol \) components are used as the second view.

Second, we want to analyze the possibilities of using multispectral images for the training of CMC models. To accomplish this we train our CMC models on subsets of 10,000 (10k), 25,000 (25k), 50,000 (50k), 100,000 (100k), and 196,215 (200k) images from BigEarthNet dataset and use them as feature extractors on BigEarthNet and So2Sat downstream tasks. We propose two possible ways to create multispectral image views for the CMC model. Our first method is based on the idea that the first view should contain the spectral bands with short wavelengths, while the second view should contain the ones with long wavelengths. In addition, we also split visible bands into different views so that the blue color channel ends up in the view with long-wavelength bands, while red and green end up in the other view. Views created in this way are shown in Table~\ref{tab:defined_view} where we used only 10 spectral bands available from Sentinel-2 as proposed in~\cite{sumbul2020bigearthnet}. The second method for defining CMC views is based on the idea that it is better to use decorrelated channels when training CMC models~\cite{tian2019contrastive,tian2020makes}. To decorrelate previously used spectral bands we employ Principal Component Analysis - PCA. To determine the PCA decomposition we randomly sampled 144 pixel values from each of 200k training images from BigEarthNet. After determining the PCA decomposition, original image channels are projected into the PCA basis and new CMC views are defined so that they contain approximately the same variance. To be precise, the first view contains the channel that explains the most data variance and four channels that explain the least amount of variance, while the second view contains the remaining channels.

\begin{table}
    \centering
    \begin{tabular}{cc}
         \toprule
         View & Spectral bands  \\
         \midrule
         1       & 2, 8, 8A, 11, 12 \\
         2  & 3, 4, 5, 6, 7  \\
         \bottomrule
    \end{tabular}
    \vspace{0.3em}
    \caption{One possible method for defining CMC views of multispectral images.}
    \label{tab:defined_view}
\end{table}

\begin{figure*}[ht]
   \centering
   \includegraphics[width=0.95\linewidth]{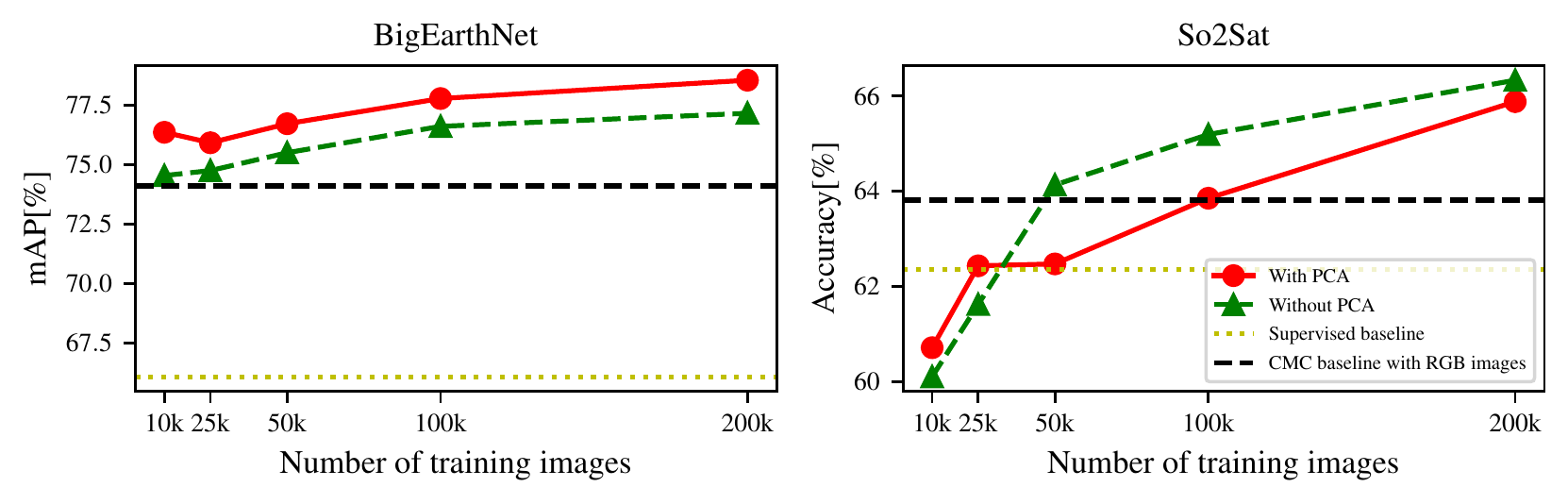}
   \caption{Results obtained using a linear classifier on top of the representations learned by self-supervised learning on multispectral images. Supervised baseline represents a baseline obtained using supervisedly pre-trained network on ImageNet, and CMC baseline with RGB images represents the best result obtained in previous experiments.}
   \label{fig:multispectral}
\end{figure*}

Finally, we evaluated the possibility of using previously trained CMC models for finetuning on our downstream tasks. For this, we used only CMC models trained using 200k images from different datasets and finetuned them using the procedure presented in Section~\ref{sec:transfer}.

All experiments are implemented on a computer with two NVIDIA GeForce RTX 2080Ti GPUs and using a Pytorch~\cite{paszke2019pytorch} library. Depending on the number and type (RGB or multispectral) of images used for training of the CMC models, training a model takes from few hours to few days. As for the transfer learning, all feature extraction models can be trained in a couple of minutes, while finetuning can take up to a few days in the case of multispectral images.

\section{Experimental Results}

\subsection{Feature extraction with RGB images}

\newcolumntype{L}{>{\centering\arraybackslash}m{1.3cm}}
\newcolumntype{R}{>{\centering\arraybackslash}m{2.0cm}}
\newcolumntype{K}{>{\centering\arraybackslash}m{2.1cm}}
\begin{table*}[ht]
    \centering
    \begin{tabular}{cKKLLRLL}
         \toprule
         & \multicolumn{7}{c}{Finetuning initialization} \\ \cmidrule{2-8}
         & \multicolumn{5}{c}{RGB images} & \multicolumn{2}{c}{Multispectral images} \\\cmidrule(lr){2-6} \cmidrule(lr){7-8}
         \multirow{-3}{*}{Downstream task}& NWPU-DOTA HR 200k & BigEarthNet LR 200k & ImageNet 200k & ImageNet 1.2M & ImageNet supervised & Without PCA & With PCA \\
         \midrule
         AID (HR)       & 94.76 & 94.60 & 94.60 & 94.16 & \textbf{95.58} & - & - \\
         MLRSNet (HR)  & 95.49 & 95.02 & 95.31 & 92.59 & \textbf{96.43} & - & - \\
         BigEarthNet (LR) & 78.34 & 79.56 & 78.63 & 76.21 & 81.62 & 82.00 & \textbf{82.90}  \\
         So2Sat (LR) & 58.43 & 58.50 & 58.51 & 57.06 & \textbf{60.84} & 60.66 & 60.68 \\
         \bottomrule
    \end{tabular}
    \vspace{0.3em}
    \caption{Finetuning results on our downstream tasks. Abbreviations HR and LR represent high-resolution and low-resolution remote sensing datasets, respectively.}
    \label{tab:finetuning_results}
\end{table*}

In Figure~\ref{fig:linear} the results obtained using a linear classifier on top of the representations learned by CMC models trained on different datasets with varying sizes are shown. In the same Figure, we also show the baseline results obtained by using the representations from the CMC model trained using 1.2M ImageNet images and ResNet50 model trained in supervised fashion on ImageNet dataset, available in PyTorch~\cite{paszke2019pytorch}. As we can see, using bigger datasets for the training of CMC models, in general, gives better representations for our downstream tasks, but this can be expected from most deep learning models. More interesting is to look at the influence of the domains of images used for CMC training on the results on downstream tasks. For downstream tasks on AID and MLRSNet datasets, we can see that using NWPU-DOTA dataset for self-supervised pre-training gives better results than using ImageNet or BigEarthNet for pre-training while using ImageNet gives better results than using BigEarthNet. Interestingly, using even as low as 25k or 50k images from NWPU-DOTA gives better results than using 200k images from ImageNet and in the case of MLRSNet even better than using 1.2M images from ImageNet. Besides that, using NWPU-DOTA dataset for self-supervised pre-training can give results that are better than the results obtained using representations learned by supervised learning on ImageNet. In the case of testing on BigEarthNet and So2Sat datasets, it is visible that using BigEarthNet dataset for self-supervised pre-training gives better results than using NWPU-DOTA or ImageNet for pre-training while using NWPU-DOTA gives better results than using ImageNet. For the case of testing on BigEarthNet, we can see that using even 10k images from BigEarthNet for self-supervised pre-training gives results that are better than any other result obtained either using supervised or self-supervised learning. As for the So2Sat, we can see that using 25k or more images from BigEarthNet, for self-supervised learning, gives results that are better than the results obtained using self-supervised learning on other datasets. Besides that, using 50k images from BigEarthNet for self-supervised learning exceeds the results obtained by using the supervisedly pre-trained network.

The aforementioned results show us that the choice of images used for the training of the CMC model is crucial to the performance on the downstream tasks if we use CMC models as feature extractors. We can see that on the downstream task of remote sensing image classification it is much better to train the CMC model on remote sensing images than on images of natural scenes. This is even more pronounced when the spatial resolutions of the images used in the pretext and downstream tasks are similar. Interestingly, if the spatial resolution between pre-training and downstream images significantly differs we get some opposed results. For the case of high-resolution downstream tasks, we get better results by pre-training on images of natural scenes than on low-resolution remote sensing images. On the other hand, when we do testing on low-resolution tasks it is better to pre-train on high-resolution remote sensing images than on images of natural scenes. We think that this discrepancy might be due to the inability of learning sharp edge detection filters when we pre-train on low-resolution images. On the other hand, if we pre-train on high-resolution remote sensing images we can learn both filters for sharp edge detection and filters that would detect features interesting for the special case of remote sensing images.

\subsection{Feature extraction with multispectral images}

In Figure~\ref{fig:multispectral}, we can see the results obtained using a linear classifier on top of the representations learned by self-supervised learning on multispectral images. As we would have expected, using bigger datasets for self-supervised pre-training gives better results on downstream tasks. Besides, we can see that on the BigEarthNet downstream task using PCA-based views turns out to be better by around 2\% than using non-decorrelated views. On the other hand, testing on the So2Sat dataset shows that better results are obtained if we use non-decorrelated views, but differences in results are in general much smaller. We think that this difference might come from the fact that the PCA decomposition was learned using images from the BigEarthNet and the learned basis may be too specialized to that dataset.

If we compare the results obtained using multispectral images with the ones obtained using only RGB images we can see that using multispectral information boosted the results by 2\% to 4\%.

\subsection{Finetuning}

Results obtained using finetuning are shown in Table~\ref{tab:finetuning_results}. We can see that in the case of RGB images the best results are obtained using neural networks pre-trained in a supervised fashion. These models show around 1\% to 2\% better results than the models pre-trained in a self-supervised manner. It is also interesting that all self-supervised models give almost the same results with only a slight favor to the models trained with images that have similar resolution as the images in the downstream task. When we use multispectral images, finetuning a self-supervised network will give around 1\% better results on the BigEarthNet dataset compared to the supervisedly pre-trained network and almost the same results on the So2Sat. However, it is important to note that on the So2Sat finetuned models give worse results than linear classifiers. We have observed a significant overfitting problem on the So2Sat using both linear and finetuned models that we were not able to solve using architectures and hyperparameters used in this work.

\section{Conclusion}
In this paper, we evaluate the applicability of self-supervised learning in remote sensing image classification. We show that if we use pre-trained models as feature extractors, self-supervised models trained on RGB remote sensing images give better results than supervised models trained on ImageNet. This can be seen even if the datasets used for self-supervised pre-training are 6 to 24 times smaller than the ImageNet dataset. However, if finetuning is used, the models pre-trained in a supervised fashion on ImageNet give up to 2\% better results than the models trained using self-supervised learning. Nevertheless, self-supervised pre-training still has merit because finetuning of the network is time and resource consuming. Furthermore, models pre-trained on multispectral images are rare and we showed that self-supervised pre-training on multispectral images boosts the performance on downstream tasks up to 4\% and allows us to match or surpass the results obtained by finetuning a supervised model pre-trained on RGB images.

There are still a lot of directions for future research in this area. In this paper we focused on scene classification, but it would be important to investigate whether similar results would be obtained for different downstream tasks, such as segmentation or object detection. Next, new methods for creating views for multispectral images are needed, since PCA-based views fail to generalize to different downstream tasks. It would also be interesting to train self-supervised methods on huge amounts of unlabeled data to find the limits of these methods. Finally, further investigation why supervisedly trained models give better results with finetuning is necessary.

\paragraph{Acknowledgements}
This research was funded in part by the Ministry of Scientific and Technological Development, Higher Education and Information Society of the Republic of Srpska under contract 19.032/961-102/19, “Service for Classification of Remote Sensing Images”.

{\small
\bibliographystyle{ieee_fullname}
\bibliography{egbib}
}

\end{document}